\documentclass[11pt,letterpaper]{article}
\usepackage{cogsys}
\usepackage[T1]{fontenc}
\usepackage{times}
\usepackage[pdftex]{graphicx} % use this when importing PDF files

\usepackage{natbib}

\usepackage{times}
\usepackage{soul}
\usepackage{url}
\usepackage[hidelinks]{hyperref}
\usepackage[utf8]{inputenc}
\usepackage{graphicx}
\usepackage{amsmath}
\usepackage{amsthm}
\usepackage{booktabs}
\usepackage{algorithm}
\usepackage{algorithmic}
\usepackage{xspace}
\usepackage{booktabs}
\usepackage{makecell}
\usepackage{array,multirow,graphicx}
\newcommand{\code}[1]{\textbf{#1}}

\newcommand{\rel}[3]{$\langle #1,\code{#2},#3 \rangle$}

\newcommand{\hide}[1]{}

\newcommand{\spear}{SpEAR\xspace}
\newcommand{\spert}{SpERT\xspace}

\newcommand{\sciclaim}{SciClaim\xspace}

\newcommand{\secref}[1]{Section~\ref{sec:#1}}

\newcommand{\figref}[1]{Figure~\ref{fig:#1}}

\newcommand{\tabref}[1]{Table~\ref{tab:#1}}

\newcommand{\textq}[1]{``#1''}

\cogsysheading{11}{2021}{1--**}{9/2021}{11/2021}
\ShortHeadings{Causal Knowledge Graph Extraction}
              {S.\ Friedman, I.\ Magnusson, V. Sarathy, and S.\ Schmer-Galunder}

\begin{document}

\title{From Unstructured Text to Causal Knowledge Graphs: \\ A Transformer-Based Approach}

\author{Scott Friedman}{friedman@sift.net}
\author{Ian Magnusson}{imagnusson@sift.net}
\author{Vasanth Sarathy}{vsarathy@sift.net}
\author{Sonja Schmer-Galunder}{sgalunder@sift.net}
\address{SIFT, 319 N 1st Ave.,
         Minneapolis, MN 55401 USA}
\vskip 0.2in

\newtheorem{example}{Example}
\newtheorem{theorem}{Theorem}

\begin{abstract}

Qualitative causal relationships compactly express the direction, dependency, temporal constraints, and monotonicity constraints of discrete or continuous interactions in the world.
In everyday or academic language, we may express interactions between quantities (e.g., sleep decreases stress), between discrete events or entities (e.g., a protein inhibits another protein's transcription), or between intentional or functional factors (e.g., hospital patients pray to relieve their pain).
Extracting and representing these diverse causal relations are critical for cognitive systems that operate in domains spanning from scientific discovery to social science.
% These causal relations are not complete mechanism descriptions in themselves, but we use them frequently in everyday language and formal instruction to express causality, allowing us to hedge when numerical details and other constraints are uncertain.
This paper presents a transformer-based NLP architecture that jointly extracts knowledge graphs including (1) variables or factors described in language, (2) qualitative causal relationships over these variables, (3) qualifiers and magnitudes that constrain these causal relationships, and (4) word senses to localize each extracted node within a large ontology.
We do not claim that our transformer-based architecture is itself a cognitive system; however, we provide evidence of its accurate knowledge graph extraction in real-world domains and the practicality of its resulting knowledge graphs for cognitive systems that perform graph-based reasoning.
We demonstrate this approach and include promising results in two use cases, processing textual inputs from academic publications, news articles, and social media.

\end{abstract}

% \scott{Full papers are 7p + addl 1p refs.  Short papers are 3p + addl 1p refs.}

\section{Introduction}
\label{sec:intro}

People express causal relationships in everyday language and scientific texts to capture the relationship between quantities or entities or events, compactly communicating how one event or purpose or quantity might affect another.
These causal relations are not complete mechanisms in themselves, but we use them frequently in everyday language and formal instruction to express causality, allowing us to avoid unnecessary detail or to hedge when details are uncertain.

Identifying these causal relationships from natural language---and also properly identifying the factors that they relate---remains a challenge for cognitive systems.
This difficulty is due in part to the expressiveness of our language, e.g., the multitude of ways we may describe how an experimental group scored higher on an outcome than a control group, and also due to the complexity of the systems we describe.

This paper describes an approach to automatically extracting (1) entities that are the subject of causal relationships, (2) causal relationships describing mechanisms, intentions, monotonicity, and temporal priority, (3) multi-label attributes to further characterize the causal structure, and (4) ontologically-grounded word senses for applicable nodes in the causal graph.
Our primary claim is that context-sensitive language models can detect and characterize the qualitative causal structure of everyday and scientific language in a representation that is usable by cognitive systems.
As evidence, we present our \spear transformer-based NLP model based on BERT \citep{devlin-etal-2019-bert} and \spert \citep{eberts2019span} that extracts causal structure from text as knowledge graphs, and we present promising initial results on (1) characterizing scientific claims and (2) representing and traversing descriptive mental models from ethnographic texts.

The present work aims to infer causal, functional, and intentional relational structure, so its output knowledge representations are relevant to cognitive systems; however, the NLP methodology that performs the inference is not intended to model human cognition.
The nodes within the causal, semantic graphs produced by \spear link to the WordNet word sense hierarchy \citep{fellbaum2010wordnet} to facilitate subsequent reasoning.
Unlike rule-based parsers that use ontological constraints during the parsing process, our NLP architecture infers ontological labels (i.e., WordNet word senses) as a context-sensitive post-process.
We demonstrate that the knowledge representations inferred by our system allows traversal across concepts to characterize meaningful causal influences.

We continue with a review of related work in qualitative causal representations (\secref{qr}) and transformer-based NLP (\secref{nlp}).
We then describe our approach (\secref{approach}) and preliminary results in two domains (\secref{results}).
We conclude with a discussion of future work in this area.

\begin{figure*}[htb]
\centering
\includegraphics[width=\linewidth]{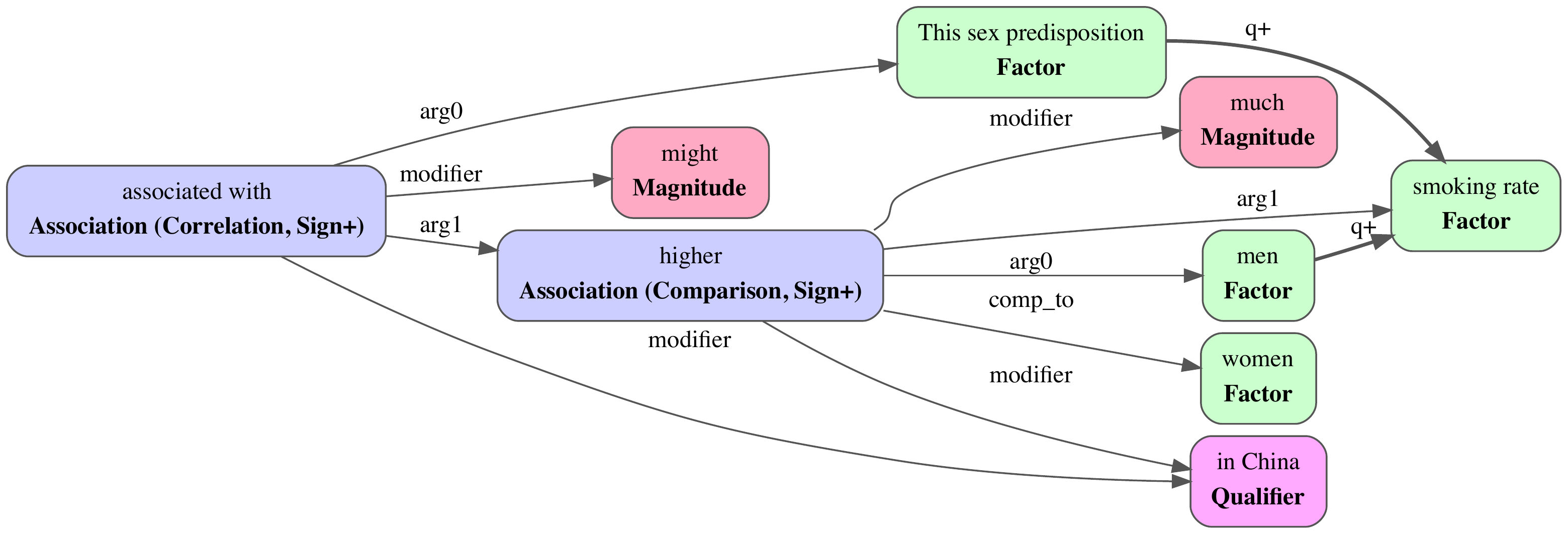}
\vspace{-0.1in}
\caption{\spear knowledge graph output for the text \textq{This sex predisposition might be associated with the much higher smoking rate in men than in women in China.} Includes a correlation, a comparison with a qualitative increase, magnitudes, and a location qualifier.}
\label{fig:score1}
\vspace{0.3in}
\includegraphics[width=.8\linewidth]{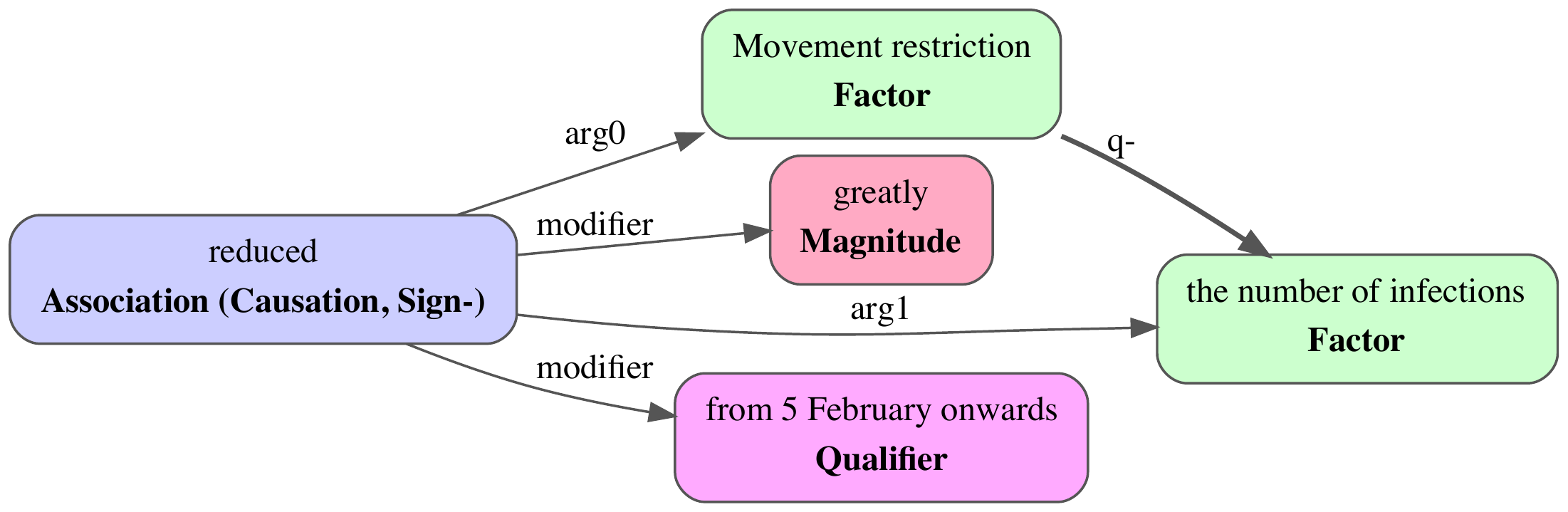}
\caption{\spear knowledge graph output for \textq{Movement restriction greatly reduced the number of infections from 5 February onwards.} Includes a causal association, a qualitative decrease, a magnitude, and a temporal qualifier.}
\label{fig:score2}
\end{figure*}

\section{Background and Related Work}

We review related work in representing causal relations, which informs the present approach.
We then review previous work in transformer-based NLP, including the \spert system \citep{eberts2019span} which is a subsystem of our architecture.

\subsection{Qualitative Causal Relations}
\label{sec:qr}

The knowledge representations described in this paper are motivated by previous work in qualitative reasoning and simulation \citep{forbus2019qualitative}.
For example, \emph{qualitative proportionalities} describe how one quantity impacts another, in a directional, monotonic fashion.
In this work, we designate \rel{a}{q+}{b} (and respectively, \rel{a}{q–}{b}) as qualitative proportionalities from $a$ to $b$, such that increasing $a$ would increase (and respectively, decrease) $b$.
This is motivated by quantity-to-quantity $\alpha_{Q+/-}$ relations \citep{forbus1984qualitative} and $M^{+/-}$ relations in qualitative simulation \citep{kuipers1986qualitative}.  
Our semantics are less constrained than either of these, due to tendencies in language to express an increase from an event to a quantity (e.g., \textq{smoking a cigarette will increase your risk of cancer}) or from entities to activities (e.g., \textq{the prime increased participants' retrieval of the cue}), and so on.

Previous work in philosophy \citep{dennett1989intentional} and cognitive psychology \citep{lombrozo2006functional} has acknowledged intentional (i.e., psychological, goal-based) and teleological (i.e., functional, design-based) relationships as types of causal relations.
Previous work has represented these as lexical qualia or affordances \citep{pustejovsky1991syntax}.
In this work, we represent purposeful, intentional actions as a qualitative relationship \rel{a}{intent+}{b}, such that the actor of action $a$ may have intended the purpose or goal $b$, e.g., \textq{they prayed for a safe pregnancy.}
We represent teleological (i.e., functional or design-based) causal relations as \rel{a}{function+}{b} to indicate that the action or artifact $a$ is designed or otherwise has a function to achieve $b$, e.g., \textq{the artifacts provide protection for pregnant women.}

\subsection{Causal and Transformer-Based NLP}
\label{sec:nlp}

Transformer-based methods for NLP utilize neural networks to encode a sequence of textual tokens (i.e., words or sub-words) into large vector-based representations for each token, sensitive to the context of the surrounding tokens \citep{devlin-etal-2019-bert}.
This is widely regarded as a state-of-the-art methodology for NLP, and has been used to process text to extract knowledge graphs, e.g., of people and relations \citep{eberts2019span}.
The architecture we present in this paper has been applied to the \sciclaim dataset of scientific claims \citep{sciclaim2021} and a social media corpus centered on moral attributions and hate speech \citep{tybalt2021cogsci}.
Many existing transformer models---similar to the architecture presented in this paper---require hundreds (sometimes thousands) of labeled training examples to reach high proficiency.

Existing symbolic semantic parsers extract scientific claims and assertions from text with explicit relational knowledge representations \citep{allen2015complex}. 
Many of these rely on rule-based engines with hand tuning, which provides more customization and interpretability, at the expense of using NLP experts to maintain and adapt to new domains.
By contrast, our approach extracts causal knowledge graphs using advances in transformer-based models such as \spert \citep{eberts2019span} to learn graph-based representations from examples alone.
The resulting knowledge graphs are ontologically-grounded and support graph-based reasoning, as we demonstrate below; however, these are not as expressive as some modern symbolic parsers.

Other NLP approaches use machine learning to extract features from scientific texts, e.g., to identify factors and directions of influence in scientific claims \citep{mueller2019deepcause}; however these approaches do not explicitly infer relations between elements in a causal graph or the ontological groundings of the terms, as in our approach.

\section{Approach}
\label{sec:approach}

We describe our graph schema for representing the entities, attributes, and qualitative relationships extracted from text.  
We discuss the general problem definition and then we explain the specific graph schemas in two domains: (1) scientific claims and (2) ethnographic mental models.  

\subsection{Knowledge Graphs}

The \spear knowledge graph format includes the following three types of elements: entities, attributes, and relations.
We describe each of these before defining the problem and describing the architecture.

\begin{figure}[htb]
\centering
\includegraphics[width=\linewidth]{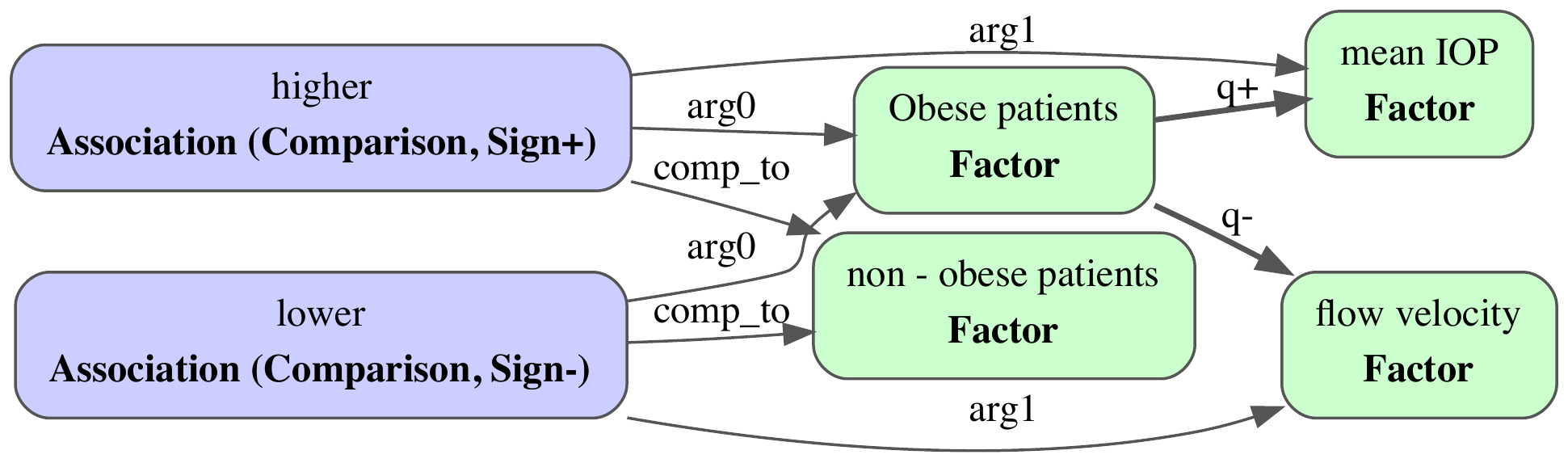}
\caption{\spear knowledge graph output for \textq{Obese patients have a higher mean IOP and lower flow velocity than non-obese patients.} The two qualitative comparisons \textq{higher} and \textq{lower} support qualitative \code{Sign+} and \code{Sign-} attributes, and \code{q+} and \code{q-} relations, respectively.}
\label{fig:score3}
\end{figure}

\paragraph{Entities.}
Entities are labeled spans within a textual example.
These are the nodes in the knowledge graph.
The same exact span cannot correspond to more than one entity type, but two entity spans can overlap.
Entities comprise the nodes of Figures \ref{fig:score1}-\ref{fig:score3} upon which attributes and relations are asserted.
Unlike most ontologically-grounded symbolic parsers \citep[e.g.,][]{das2010probabilistic,allen2015complex}, these entity nodes are not ontologically grounded in a class hierarchy; rather, these entity nodes are associated with a token sequence (e.g., \textq{smoking rate} in \figref{score1}) and a corresponding entity class (e.g., \code{Factor}).
These entities also have high-dimensional vectors from the transformer model, which approximates the distributed semantics.
Our architecture also associates entities with applicable WordNet senses, as we describe below in \secref{wordnet}.

\paragraph{Attributes.}
Attributes are Boolean labels, and each entity (i.e., graph node) may have zero or more associated attributes. 
Attribute inference is therefore a multi-label classification problem.
The previous \spert transformer model was not capable of expressing these; this is a novel contribution of \spear, as described in \secref{arch}.
In Figures \ref{fig:score1}-\ref{fig:score3}, attributes are rendered as parenthetical labels inside the nodes, e.g., \code{Correlation} and \code{Sign+} in the \figref{score1} nodes for \textq{associated with} and \textq{higher,} respectively.
The multi-label nature allows the \figref{score1} \textq{higher} node to be categorized simultaneously as \code{Sign+} and \code{Comparison}.

\paragraph{Relations.}
Relations are directed edges between labeled entities, representing semantic relationships.
These are critical for expressing what-goes-with-what over the set of entities.
For example in the sentence in \figref{score1}, the relations (i.e., edges) indicate that the \textq{higher} association asserts the antecedent (\code{arg0}) \textq{men} against (\code{comp\_to}) \textq{women} for the consequent (\code{arg1}) \textq{smoking rate.}
In Figures \ref{fig:score1}-\ref{fig:score3}, the \code{modifier} relations link nodes to others that semantically modify them.
Without all of these labeled relations, the semantic structure of these scientific claims would be ambiguous.

\subsection{Problem Definition}

We define the multi-attribute knowledge graph extraction task as follows: for a text passage $\mathcal{S}$ of $n$ tokens $s_1,...,s_n$, and a schema of entity types $\mathcal{T}_e$, attribute types $\mathcal{T}_a$, and relation types $\mathcal{T}_r$, predict:
\begin{enumerate}
    \item The set of entities $\langle s_j, s_k, t \in \mathcal{T}_e \rangle \in \mathcal{E}$ ranging from tokens $s_j$ to $s_k$, where $0 \leq j \leq k \leq n$,
    \item The set of relations over entities $\langle e_{head} \in \mathcal{E}, e_{tail} \in \mathcal{E}, t \in \mathcal{T}_r \rangle \in \mathcal{R}$ where $e_{head} \neq e_{tail}$,
    \item The set of attributes over entities $\langle e \in \mathcal{E}, t \in \mathcal{T}_a \rangle \in \mathcal{A}$.
\end{enumerate}
This defines a directed multi-graph without self-cycles, where each node has zero to $|\mathcal{T}_a|$ attributes. 
\spear does not presently populate attributes on relations.

\subsection{Knowledge Graph Schemas}

We briefly describe a subset of the graph schemas for our two use-cases: (1) the \sciclaim dataset of scientific claims and (2) ethnographic mental models.
These two schemas share some qualitative causal representations but vary in other domain-specific descriptions. 
In follow-on work, these schemas may be integrated into a single schema.

\paragraph{Scientific Claims.}
The \sciclaim scientific claim schema is designed to capture associations between factors (e.g., causation, comparison, prediction, proportionality), monotonicity constraints across factors, epistemic status, and high-level qualifiers.
This model is used for qualitative reasoning to help characterize the replicability and reproducibility of scientific claims \citep{alipourfard2021systematizing,gelman2021toward}.
We describe the entities, attributes, and relations of the schema, referencing the graphed examples rendered by our system in Figures \ref{fig:score1}, \ref{fig:score2}, and \ref{fig:score3}.

This schema includes six entity types:
\textbf{Factors} are variables that are tested or asserted within a claim (e.g., \textq{smoking rate} in \figref{score1});
\textbf{Associations} are explicit phrases associating one or more factors in a causal, comparative, predictive, or proportional assertion (e.g., \textq{associated with} and \textq{reduced} in Figures \ref{fig:score1} and \ref{fig:score2}, respectively);
\textbf{Magnitudes} are modifiers of an association indicating its likelihood, strength, or direction (e.g., \textq{might} and \textq{much} in \figref{score1});
\textbf{Evidence} is an explicit mention of a study, theory, or methodology supporting an association;
\textbf{Epistemics} express the belief status of an association, often indicating whether something is hypothesized, assumed, or observed;
\textbf{Qualifiers} constrain the applicability or scope of an assertion (e.g., \textq{in China} in \figref{score1} and \textq{from 5 February onwards} in \figref{score2}).

This schema includes the following attributes, all of which apply solely to the \textit{Association} entities:
\textbf{Causation} expresses cause-and-effect over its constituent factors (e.g., \textq{reduced} span in \figref{score2});
\textbf{Comparison} expresses an association with a frame of reference, as in the \textq{higher} statement of \figref{score1} and the \textq{higher} and \textq{lower} statements of \figref{score3};
\textbf{Sign+} expresses high or increased factor value;
\textbf{Sign-} expresses low or decreased factor value;
\textbf{Indicates} expresses a predictive relationship; and \textbf{Test} indicates a statistical test employed to test a hypothesis.

We encode six relations:
\textbf{arg0} relates an association to its cause, antecedent, subject, or independent variable;
\textbf{arg1} relates an association to its result or dependent variable;
\textbf{comp\_to} is a frame of reference in a comparative association;
\textbf{modifier} relates entities to descriptive elements;
\textbf{q+} and \textbf{q-} indicate positive and negative qualitative proportionality, respectively, where increasing the head factor increases or decreases (the amount or likelihood of) the tail factor, respectively.

\begin{figure*}[tb]
\centering
\includegraphics[width=\linewidth]{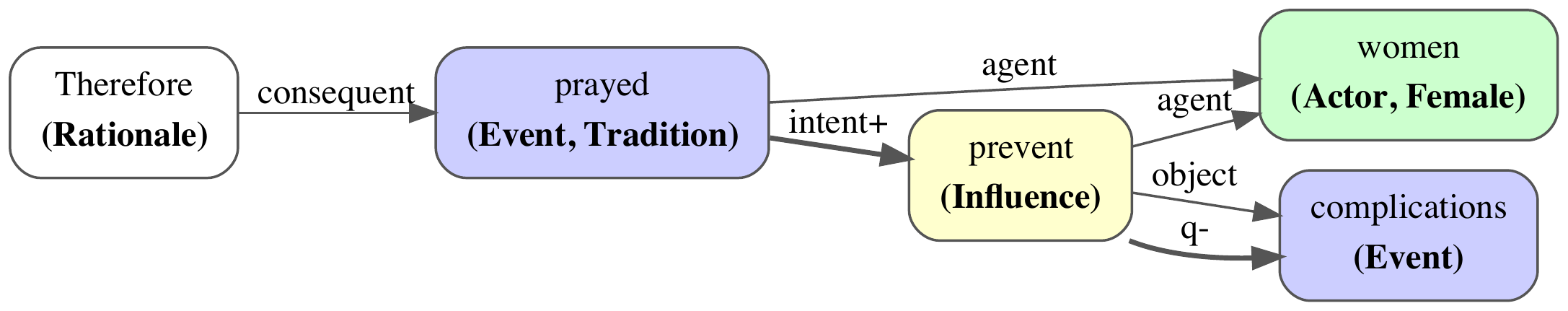}
\vspace{-0.1in}
\caption{\spear knowledge graph for \textq{Therefore, the women prayed to prevent any complications,} including \code{intent+} and \code{q-} relations.}
\label{fig:hab1}
\vspace{0.3in}
\includegraphics[width=.8\linewidth]{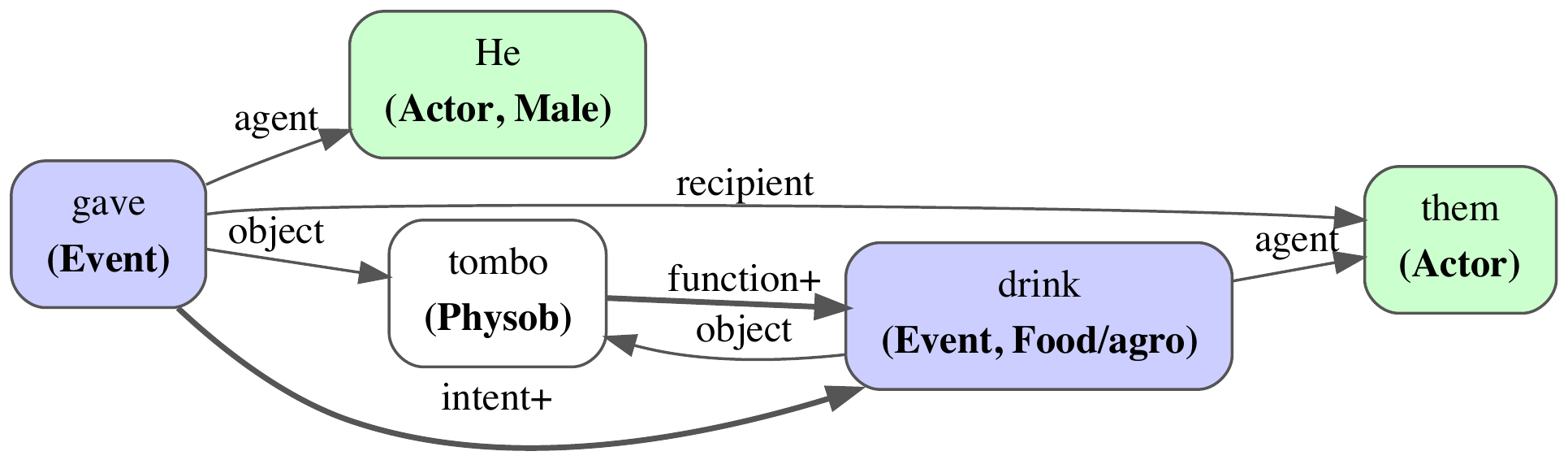}
\caption{\spear knowledge graph for \textq{He also gave them tombo to drink.}, including \code{intent+} and \code{function+} relations.}
\label{fig:hab2}
\end{figure*}

\paragraph{Ethnographic Mental Models.}

In our preliminary ethnographic mental modeling domain, we utilize a slightly different schema to capture intentional and functional causality in addition to culturally-specific attributes such as gender and spirituality.
We use \figref{hab1} and \figref{hab2} to illustrate the ethnographic causal graph schema.

This schema includes attributes for spiritual or cultural \code{Tradition} (e.g., \textq{prayed} in \figref{hab1}), \code{Event} (e.g., \textq{gave} and \textq{drink} in \figref{hab2}), \code{Influence} for causally-potent elements (e.g., \textq{prevent} in \figref{hab1}), and others.

We include additional relations \code{agent}, \code{object}, \code{recipient}, \code{consequent}, and others as semantic role relations of events and assertions.
These relations (rendered in narrow lines in \figref{hab1} and \figref{hab2}) comprise a description logic of their head nodes, such that the head node would not have the same semantics without the its reachable subgraph along these edges.

The bold-rendered edges are causal edge, including qualitative monotonicity \code{q+} and \code{q-}, temporal precedence \code{t+} relations to indicate one event preceding another, and intentional \code{intent+} and functional \code{function+} relations to indicate the goal (i.e., intention or function, respectively) of an action or artifact. 
For instance, the graph in \figref{hab1} shows an \code{intent+} from \textq{prayed} to \textq{prevent} and then a \code{q-} to \textq{complications}, ultimately indicating that prayer has a goal of minimizing complications.
Furthermore, the graph in \figref{hab2} illustrates an \code{intent+} relation from \textq{gave} to \textq{drink,} indicating the giving is intended to support the drinking.
\figref{hab2} also includes a \code{function+} relation, indicating that the \textq{tombo} is designed or cultivated for drinking.

The relatively simple statement in \figref{hab1} originates from an ethnographic article \citep{aziato2016religious} that includes interview snippets, and the sentence in \figref{hab2} is from a collection of international folktales.\footnote{https://www.worldoftales.com/}
Despite their simplicity, the \spear knowledge graphs illustrate rich multi-step causality:
\figref{hab1} indicates that prayer has the purpose of reducing the incidence (or severity of) complications, and \figref{hab2} plots a simple narrative structure indicating an agent's intention to affect the actions of other agents, as well as the function of a novel entity.

\begin{figure}[tb]
\centering
\includegraphics[width=\linewidth]{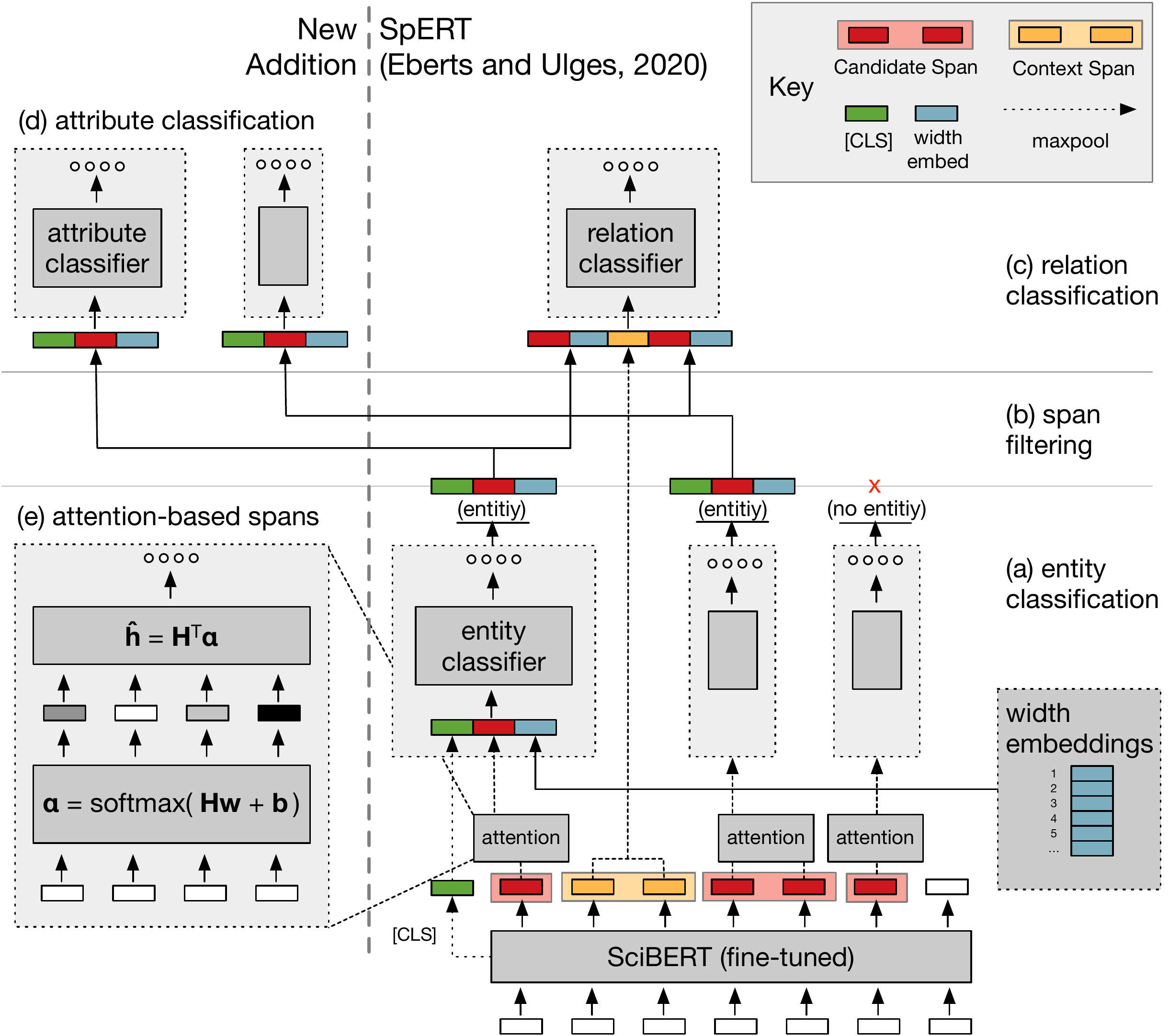}
\caption{The \spear transformer-based model extends the \spert components (a, b, and c) with attribute classification (d) that performs multi-label inference on identified entity spans and attention-based representations (e) of spans, inspired by \cite{lee-etal-2017-end}.}
\label{fig:model}
\end{figure}

\subsection{Model Architecture}
\label{sec:arch}

Our \spear model architecture extends \spert with an attribute classifier and attention-based span representation. 
The original architecture provides components (\figref{model} a--c) for joint entity and relation extraction on potentially-overlapping text spans. The parameters of the entity, attribute, and relation classifiers, as well as the parameters of the BERT language model (initialized with its pre-trained values) are all trained end-to-end on our dataset.

\subsubsection{Computing Token Vectors}

The tokens $s_1,...,s_n$ of the text passage $\mathcal{S}$ are each embedded by a transformer such as BERT \citep{devlin-etal-2019-bert} as a sequence $\textbf{e}_1,...,\textbf{e}_n$ of high-dimensional vectors representing the token and its context. BERT also provides an additional \textq{[CLS]} vector output, $\textbf{e}_0$, designed to represent information from the complete text input. 
For all possible spans,  $span_{j,k} = s_j,..., s_k$, up to a given length, the word vectors associated with a span, $\textbf{e}_j,..., \textbf{e}_k$, are combined into a final span vector, $\textbf{e}(span_{j,k})$.

\subsubsection{Computing Span Vectors}

The original \spert architecture uses \textit{maxpooling} to compute each dimension of $\textbf{e}(span_{j,k})$ as the maximum value across its constituent BERT token vectors for that dimension.
Instead of using maxpool, \spear uses an attention-based span representation (\figref{model}e) inspired by \cite{lee-etal-2017-end} to compute span vectors. 
This produces \textit{attention weight} scalars $\alpha_{i,t}$ for each BERT token vector $\mathbf{h}_t$ in a span $i$ using learned parameters $\mathbf{w}$ and $b$:
% \begin{align}
%     \alpha_t = \mathbf{w} \cdot \mathbf{h}_t + b
% \end{align}
\begin{align}
    \alpha_{i,t} = \frac{\exp(\mathbf{w} \cdot \mathbf{h}_t + b)}{\sum_{k=START(1)}^{END(i)} \exp(\mathbf{w} \cdot \mathbf{h}_k + b)}
\end{align}

\noindent
These attention weights help compute the span representation $\mathbf{\hat{h}}_i$ with the following weighted sum:
\begin{align}
    \mathbf{\hat{h}}_i = \sum_{t=START(1)}^{END(i)}\alpha_{i,t} \mathbf{h}_t
\end{align}

\subsubsection{Classifying Spans as Entities}

The final attention-based span representation, $\textbf{x}(span_{j,k})$ is made by concatenating together the attention representation $\textbf{e}(span_{j,k})$ and $\textbf{e}_0$ along with a width embedding, $\textbf{w}_l$, that encodes the number of words, $l$, in $span_{j,k}$. Each valid span length $l$ looks up a different vector of learned parameters, $\textbf{w}_l$.
\spear uses the concatenated $\textbf{x}(span_{j,k})$ vector to classify spans into mutually-exclusive entity types (including a \emph{null} type) using a linear classifier (\figref{model}a). 
Only spans identified as entities move on to further analysis (\figref{model}b).

\subsubsection{Inferring Multi-Class Attributes on Entities}

\spear uses its classified entities $\mathbf{x}^a$ as inputs to its attribute classifier (\figref{model}d) with weights $\mathbf{W}^a$ and bias $\mathbf{b}^a$.
A pointwise sigmoid $\sigma$ yields separate confidence scores $\mathbf{\hat{y}}^a$ for each attribute in the graph schema:
\begin{align}
    \mathbf{\hat{y}}^a = \sigma(\mathbf{W}^a \mathbf{x}^a + \mathbf{b}^a)
\end{align}
We train the attribute classifier with a binary cross entropy loss $\mathcal{L}_a$ 
% \begin{equation}
% \begin{split}
%     \mathcal{L}^a = {1 \over N|\mathcal{T}^a|} \sum_{i}^{N} \sum_{j}^{|\mathcal{T}^a|} y^a_j \cdot \log(\hat{y}^a_j) + \\
%     (1 - y^a_j) \cdot \log(1 - \hat{y}^a_j)
% \end{split}
% \end{equation}
summed with the \spert entity and relation losses, $\mathcal{L}_e$ and $\mathcal{L}_r$, for a joint loss:
\begin{align}
   \mathcal{L} = \mathcal{L}_e + \mathcal{L}_r + \mathcal{L}_a 
\end{align}

\spear takes only identified entity spans as input to the attribute classifier, as this approach provided best performance and aligns with the finding by \cite{eberts2019span} that training on downstream tasks yields best accuracy with strong negative samples of ground truth entities (i.e., teacher forcing).

\subsubsection{Inferring Labeled Relations between Entities}

\spear uses all pairings of classified entities (\figref{model}b) as inputs to its relational classifier (\figref{model}c).
\spear's relational classifier identical to \spert's: a multi-label linear classifier that takes each pair of entities (i.e., a relation head and a relation tail) and concatenates their span representations, width representations, and also the maxpool of the token vectors between the two entities.
The output of the relational classifier is zero or more labeled relations from the head entity to the tail entity.

The output of \spear's neural components comprises a \emph{directed multigraph} (i.e., a directed graph that is allowed to have multiple edges between any two nodes) without self-loops.
The multigraph may be disconnected, and may contain isolated nodes.
Each node (i.e., labeled entity) in the multigraph may have zero or more Boolean attributes.
Every entity, attribute, and relation in \spear's directed multigraphs includes a \emph{confidence score} between 0 and 1.

\subsubsection{Rectifying Results}

\spear includes a novel \textit{rectifier} component (not shown in \figref{model}) that prunes entities, attributes, and relations that are inconsistent with the constraints of the graph schema.
For example, relations might be constrained to originate or terminate at certain entity types, attributes may be constrained to certain entity types, and some attributes and relations may be mutually inconsistent.

When the rectifier detects a schema conflict, it uses \spear's confidence scores to remove lower-score elements to resolve it.
This strictly removes graph elements, so it cannot improve \spear's recall score--- and it may even reduce the recall score--- but empirically, we find the rectifier increases precision proportionately and ultimately increases \spear's F1 measure in some domains.

\begin{figure*}[h]
\centering
\includegraphics[width=\linewidth]{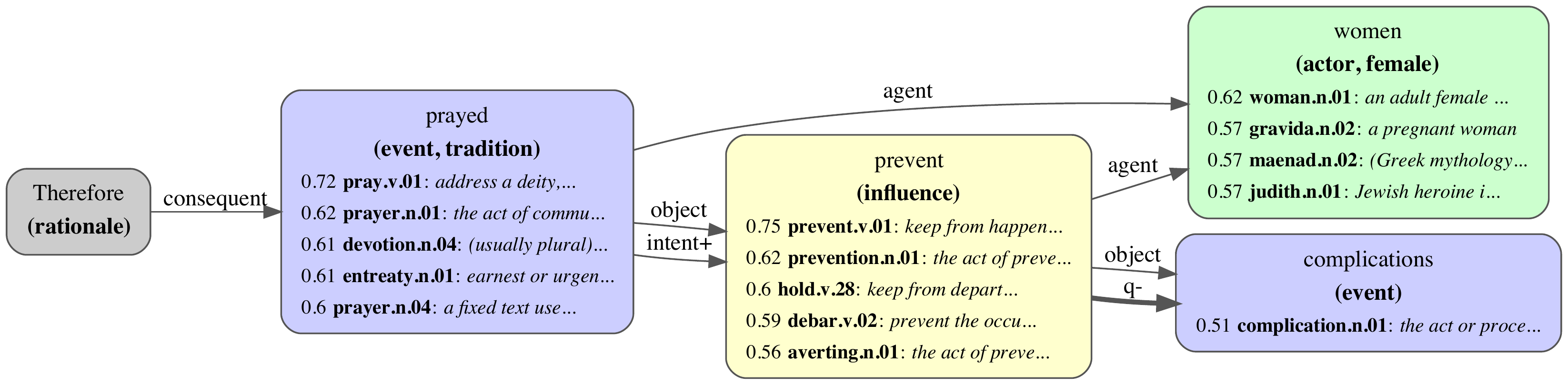}
\vspace{-0.1in}
\caption{\spear knowledge graph for the same sentence in \figref{hab1}, also displaying WordNet word senses automatically inferred by the architecture.
The listed word senses include a confidence score, the WordNet SynSet name, and a truncated WordNet definition for the inferred SynSet.}
\label{fig:chess}
\end{figure*}

\subsubsection{Inferring Word Sense}
\label{sec:wordnet}

After extracting the graph structure, our system infers a confidence distribution over word senses for each applicable node in the \spear graph, ignoring some pronouns, prepositions, determiners, and logical connectives.
\figref{chess} illustrates the output of word sense disambiguation from our system, listing all inferred word senses with a confidence score greater than 0.5.
We do not interpret the highest-confidence word sense as the single ``correct'' word sense; rather, we regard each node as having a weighted semantic locale within a lexical ontology.

Word senses are inferred using the LMMS framework \citep{loureiro-jorge-2019-language}: a transformer-based encoder encodes a vector for each token of the sentence.
Vectors for \spear nodes are computed by averaging the one or more constituent token vectors.
The system then computes the dot-product of each node's vector against pre-computed vectors for each word sense within its sense embeddings.
The dot-product results are utilized as confidence scores.

The word sense embeddings are drawn from the SynSets (i.e., synonym sets) of WordNet, a large knowledge base containing over 117,000 word senses \citep{fellbaum2010wordnet}.
Computing a confidence distribution of WordNet word senses localizes each \spear node within a structured semantic hierarchy.
This ultimately facilitates similarity-based reasoning within and across \spear graphs, e.g., by computing the least common ancestor between two different nodes within the WordNet semantic hierarchy.
These word senses are not evaluated in this paper due to lack of ground truth WordNet labels for our datasets, but word sense disambiguation is an important cognitive capability for natural language understanding, and is facilitated by the same transformer-based NLP as the rest of the architecture.

\begin{table}[htb]
\footnotesize
\centering
\begin{tabular}{c r cccc }
\toprule
& \textbf{Dimension} & \textbf{P} & \textbf{R} & \textbf{F1} & \textbf{Support} \\ \midrule
\parbox{2mm}{\multirow{7}{*}{\rotatebox[origin=c]{90}{\textbf{Entities}}}}
&factor&93.05&90.68&91.85& 2756\\
&evidence&92.17&92.00&92.04&230\\
&epistemic&91.57&73.04&81.09&299\\
&association&94.60&86.83&90.54&1290\\
&magnitude&88.19&86.76&87.46&613\\
&qualifier&78.21&78.75&78.41&360\\
& \textbf{Micro-Averaged} &  91.56 & 87.71 & 89.59 & \\
\midrule
\parbox{2mm}{\multirow{8}{*}{\rotatebox[origin=c]{90}{\textbf{Attributes}}}} 
&causation&44.64&68.00&53.85&342\\
&comparison&92.47&77.87&84.49&329\\
&indicates&85.38&70.00&76.73&84\\
&sign+&97.98&86.97&92.13&542\\
&sign-&90.22&72.14&80.13&202\\
&correlation&100.00&83.73&91.14&320\\
&test&--&--&--&25\\ % no test in the test set :(
& \textbf{Micro-Averaged} & 93.85 & 81.23 & 87.08 & \\
\midrule
\parbox{2mm}{\multirow{8}{*}{\rotatebox[origin=c]{90}{\textbf{Relations}}}} 
&arg0&84.84&75.84&80.08&1325\\
&arg1&84.74&76.69&80.50&1384\\
&comp\_to&77.92&59.20&67.27&187\\
&modifier&80.73&74.67&77.57&1582\\
&subtype&43.33&33.33&37.33&156\\
&q+&72.04&68.73&70.32&504\\
&q-&75.94&54.00&62.50&208\\
& \textbf{Micro-Averaged} & 81.37 & 73.37 & 77.16 & \\
\bottomrule
\end{tabular}
\caption{Precision, recall, F1 and support (i.e., occurrences in dataset) for \spear on the \sciclaim dataset, using 100 held-out examples from the total 901 examples in the dataset.}
\label{tab:schema-results}
\end{table}

\section{Results}
\label{sec:results}

We describe two different results of using \spear with our qualitative causal schemata: (1) precision, recall, and F1 measure in the \sciclaim scientific claims domain and (2) traversal through an ethnographic qualitative causal model.
This provides empirical evidence of the effectiveness of our approach and the expressiveness of the qualitative causal schema, respectively.

\subsection{Information Extraction for Scientific Claims}

The \sciclaim dataset for the scientific claims domain consists of 901 examples from Social and Behavior Science (SBS) literature and abstracts from PubMed and the CORD-19 dataset \citep{wang2020cord19}.
Each example consists of a single sentence labeled by a trained NLP expert with one or more spans (possibly nested) identified as entities, zero or more attributes on each entity, and zero or more relations over entities pairs (label counts are listed in \tabref{schema-results}  \textit{support}).
Most datasets for transformer-based information extraction are an order of magnitude larger.

When applying \spear to the \sciclaim dataset of scientific claims, we use the fine-tuned SciBERT transformer variant \cite{beltagy2019scibert} as the \spear input layer.

We partitioned the \sciclaim dataset into a randomized split of 100 test examples and 801 training examples, and we averaged our results over 5 train/test evaluation trials.
In each trial, we trained our \spear model for 20 epochs and then ran our evaluation.
The per-class evaluations are listed in \tabref{schema-results}, divided across the various entities, attributes, and relations. 
\tabref{schema-results} reports the micro-averaged results for entities, attributes, and relations, as well as support numbers to show the cardinality of each element in the full 901-example \sciclaim dataset.
Despite the relatively small size of the \sciclaim dataset, the model achieves promising results on most classes.
Our random train-test split included no examples of the \code{Test} attribute (which describe mentions of \textq{ANOVA,} \textq{t-test,} and other experimental methods), so \tabref{schema-results} contains no results for that row.

Importantly, the relations and attributes cannot be correct if the entities they are defined over are incorrect.
This means that we expect relations and attributes to have lower precision and recall, all else being equal.
This is especially the case for relations, which require \emph{both} of their constituent entities (i.e., head and tail nodes) to be properly characterized in order to be scored as correct.
The relations \code{q+} and \code{q-} achieved relatively low performance, due in part to the lower support in the training data, and also due to the often greater distance between these spans in the text, all else being equal.

These results across entities, attributes, and relations support our claim that qualitative causal structure can be characterized by context-sensitive NLP models.

\subsection{Extracting and Traversing Ethnography-Derived Causal Models}

In the ethnographic domain, we trained \spear on labeled examples from Anthropology papers, ethnographic manuscripts, and tweets, all related to the topic of maternal and child health in western African countries.
We then ran \spear to extract information from these and other sentences from the same types of literature.

This ethnographic dataset is roughly half the size of the \sciclaim dataset, so rather than provide another table of F1 scores, we demonstrate the reasoning capabilities supported by the NLP-extracted causal structure.
The preliminary results in \secref{valence} and \secref{traversal} include both \spear-extracted and human-labeled (i.e., ground truth) data, so we consider this a proof-of-concept study of the ability to reason over the causal models extracted by \spear.

\begin{figure*}[tb]
\centering
\includegraphics[width=.8\linewidth]{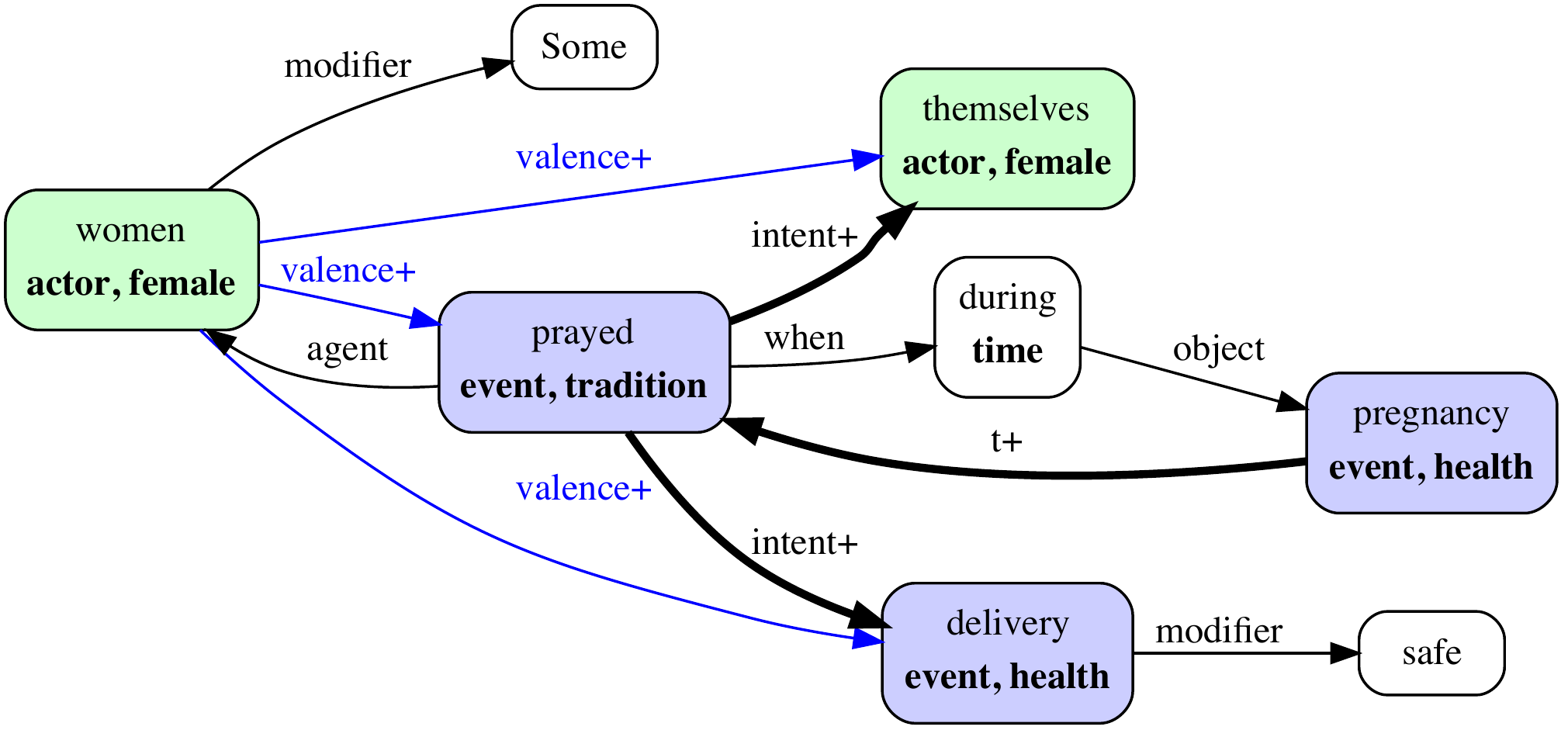}
\caption{Knowledge graph in the ethnography schema with valence inferred, for the text \textq{Some of the women prayed for themselves during pregnancy for safe delivery.}}
\label{fig:pray}
\end{figure*}

\begin{figure*}[tb]
\centering
\includegraphics[width=.9\linewidth]{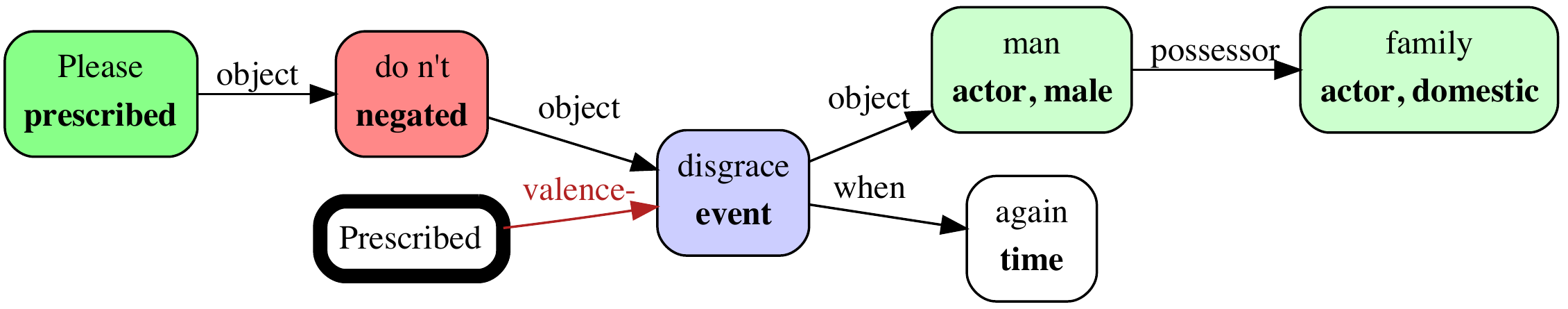}
\caption{Knowledge graph in the ethnography schema with valence inferred, for the text \textq{Please don't disgrace the man of the family again.}}
\label{fig:disgrace}
\end{figure*}

\begin{figure*}[tb]
\centering
\includegraphics[width=\linewidth]{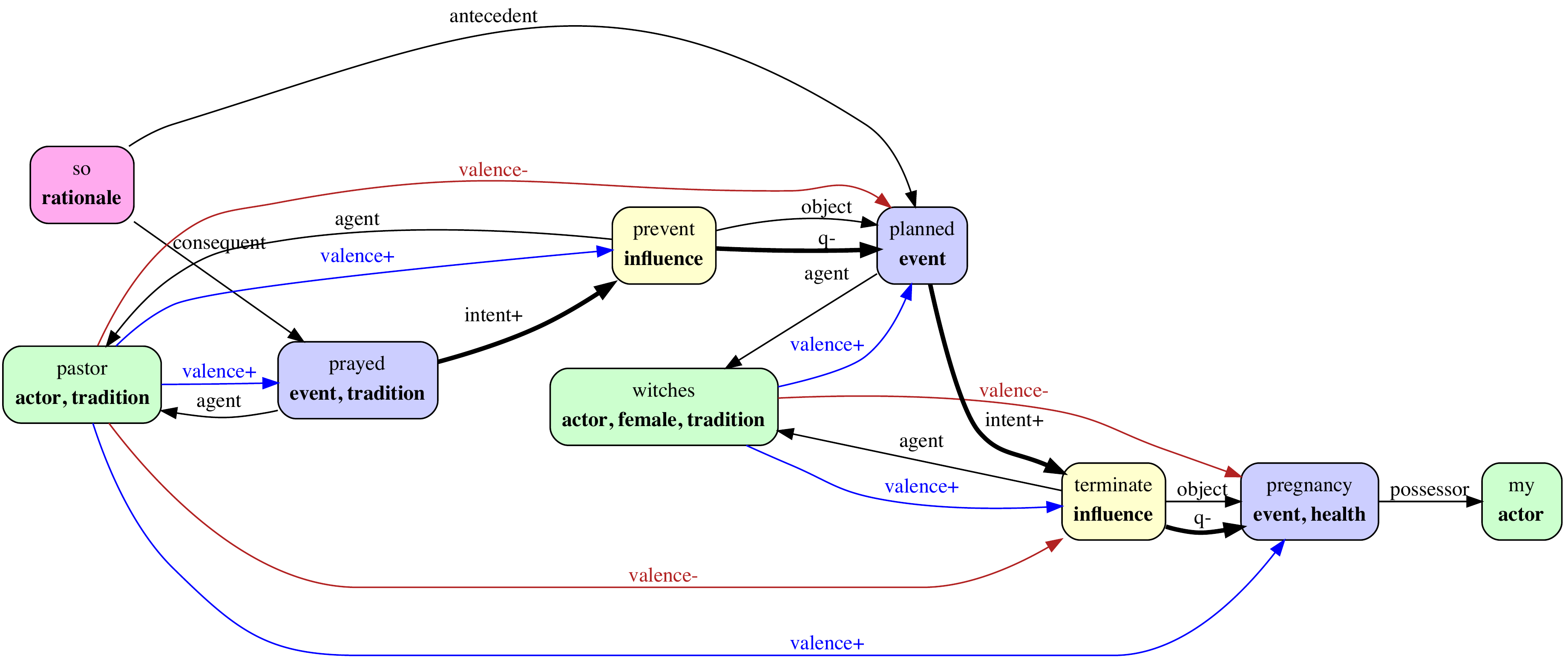}
\caption{Knowledge graph in the ethnography schema with valence inferred, for the text \textq{...the witches had planned to terminate my pregnancy, so the pastor prayed to prevent it.}}
\label{fig:witches}
\end{figure*}

\subsubsection{Computing Valence via Intentions and Qualitative Monotonicity}
\label{sec:valence}

In the ethnographic domain, the causal models in our schema (and therefore extracted by our model \spear) include intention (\code{intent+}) and function (\code{function+}) relations, which indicate an agent-based desire or normative effect of an action or artifact.

These relations indicate an agent-based or normative \emph{valence} (i.e., the positive or negative desirability) of an agent to achieve (or maximize) or prevent (or minimize) an event or quantity.
Our graph schema also includes a \code{prescribed} attribute for occurrences of \textq{should} and \textq{must} and \textq{please} that indicate a request or positive valence on behalf of the author or speaker, and a \code{negated} attribute for occurrences of \textq{don't} and \textq{not} and other negations to indicate a negation of the node's reachable subgraph.
These intentional, functional, prescribed, and negated structure in the graph support graph-based inference of agents' valence.

Valence computation starts at any \code{intent+} or \code{prescribed} source node, and then traverses forward, asserting that the \code{agent} (or otherwise a generic \code{prescribed} element) has a positive valence for that node.
When a \code{negated} attribute or \code{q-} relation is traversed, the sign of the valence is inverted.

In \figref{pray}, the traversal computes that women have positive valence for \textq{prayed} (their direct action) and \textq{themselves} (which they pray for) and \textq{safe delivery} (which they also pray for).
In \figref{disgrace}, the speaker prescribes the negation of an event, so the traversal asserts a normative negative valence for the disgracing the man of the family.
\figref{witches} is the most complex of these \spear semantic graphs, describing a pastor praying to prevent some witches' plan to terminate a pregnancy.
The traversal infers that the pastor and the witches have opposite valences toward the plan, the pregnancy termination, and the pregnancy itself.

These simple traversals over complex \spear graphs can help us infer the norms and heterogeneous values of actors across cultures, from unstructured text.

\begin{figure*}[tb]
\centering
\includegraphics[width=\linewidth]{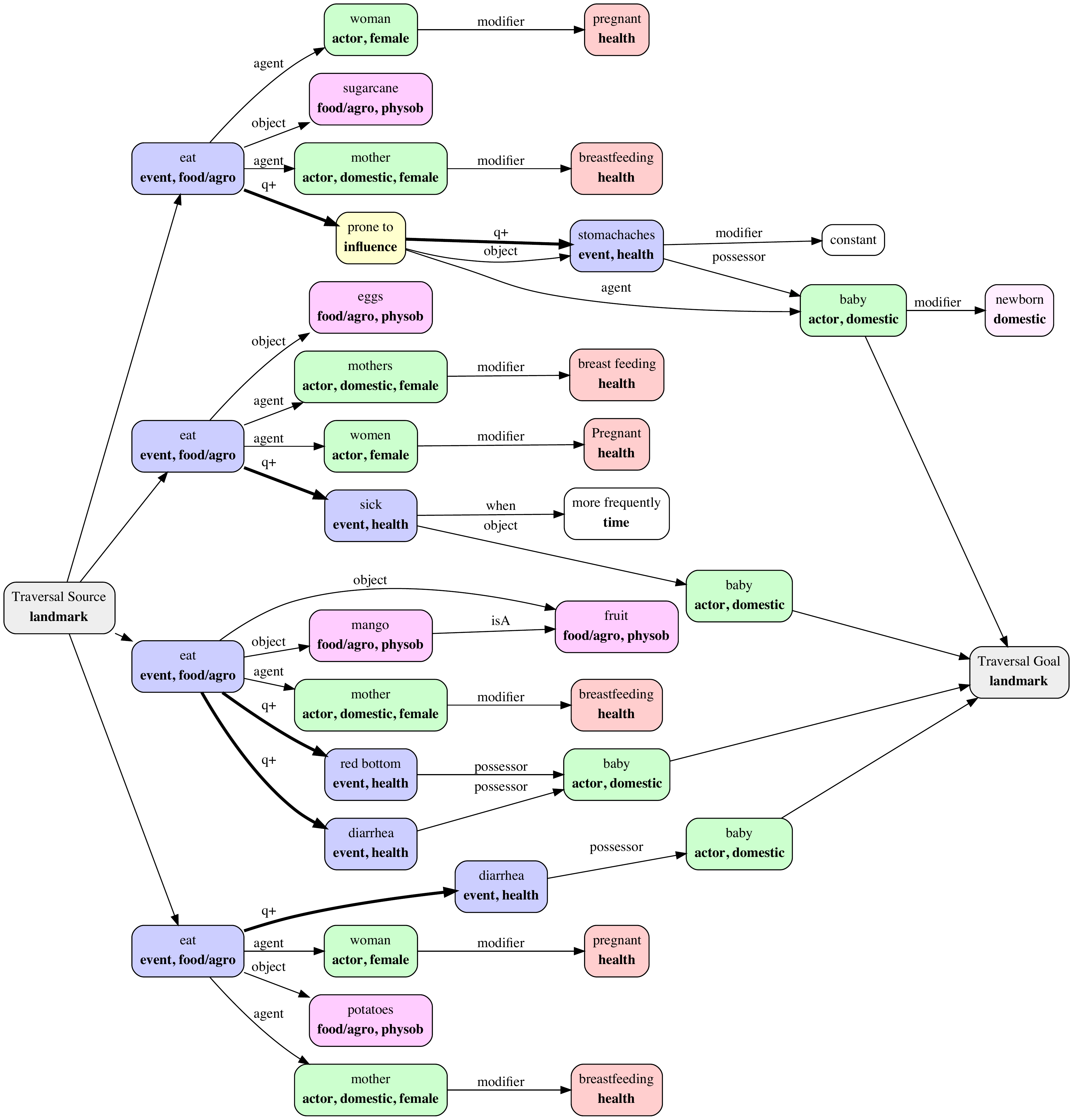}
\caption{A graph traversal from the concept \textq{eat} with an agent of lemma \textq{woman} or \textq{mother} to the concept \textq{baby} after parsing a manuscript listing common myths about maternal and child health.}
\label{fig:traversal}
\end{figure*}

\subsubsection{Finding Contextual Associations in Ethnographic Causal Models}
\label{sec:traversal}

In addition to analyzing the ethnographic models on a per-sentence basis (see above), we assemble them into a global graph comprising the \spear graphs from each sentence from the ethnographic corpus.
We implemented a simple query function that records the semantic paths between a \textit{start} and \textit{end} patterns, where the pattern could bind a node's lemma (e.g., \textq{baby}) or bind multi-node semantic graph structure, e.g., any \code{Event} with lemma \textq{eat} whose \code{agent} has lemma \textq{woman.}
The \spear nodes traversed from the query (from start to end patterns) comprise the set of relevant causal factors and relationships.

In the case of the \figref{traversal} traversal, the query begins at any \textq{eat}-lemma event performed by a \textit{mother} or \textit{woman,} and terminates at a \textq{baby}-lemma node.
Intuitively, this queries how the mother's eating might affect a baby, and \figref{traversal} iterates through eating \textq{sugarcane} \code{q+} to a baby's \textq{stomachaches,} through eating \textq{eggs} \code{q+} to a baby being \textq{sick,} through eating \textq{mango} \code{q+} to both \textq{red bottom} and \textq{diarrhea,} etc.
Note that the dietary effects extracted and traversed here are not supported by scientific evidence; rather, they are common beliefs in the regions described in the ethnography.
This query-driven traversal capability provides further evidence that the \spear causal models support practical qualitative causal reasoning.

\section{Conclusion}
\label{sec:conclusion}

This paper describes our \spear transformer-based NLP model for extracting entities, attributes, and relationships that describe qualitative causal structure.
We demonstrated the approach in the domains of (1) the \sciclaim dataset of scientific claims and (2) ethnographic corpora.
Our datasets are still under development, but despite their relative sparsity they support encouraging results with respect to F1-measure and practical reasoning capabilities via graph traversal.

One limitation of this work is that not all of the nodes generated by our approach are formally represented to support qualitative and numerical model-based reasoning.
This is due in part to the ambiguity and hedging that we see in causal language: ``smoking is associated with increased risk of lung disease'' does not unambiguously specify whether we should model ``smoking'' as a frequency, likelihood, or single occurrence, nor does it unambiguously specify whether the risk of disease increases in likelihood or severity.
The incompleteness in language---and the resulting gaps in knowledge representation---mean that some assumptions about the arguments to \code{q+} and \code{q-} may not hold in \spear's output: \code{q+} may be expressed over quantities, over events, over adjectives, or any heterogeneous mix of these, and a downstream reasoner has no formal \emph{a priori} indicator of which these are.
One remedy to this is for the NLP model to infer whether nodes are amounts, frequencies, likelihoods, etc., but it's an empirical question whether transformer-based NLP model can accurately infer these abstract categories, and it's not clear whether NLP models should attempt such inferences when the author has left it ambiguous.

Our graph traversal results suggest that the present level of representation may be adequate for use cases involving causal reasoning, graph propagation, and inferring agents' direct and indirect goals and intentions.
These are important considerations for cognitive systems that reason in scientific, causal, or social domains.

As with many modern NLP architectures, the work presented in this paper utilizes a pre-trained transformer model within its architecture.
Pre-trained transformers are trained on massive corpora collected from across the internet and other sources, which speeds up subsequent machine learning, but it also means that the sub-optimal biases of the training data---including racial, ethnic, gender, and other biases---become part of the models themselves.
Systematic biases in pre-trained models have been well-characterized \citep{garg2018word,friedman2019relating}, as have methods for de-biasing them \citep{bolukbasi2016man}; however, we note that sub-optimal biases remain a risk for any machine-learned model trained on real-world text that itself contains implicit biases.

Our near-term future work is to expand our ethnographic dataset and to utilize \spear's results in downstream systems, e.g., for estimating the reproducibility of a scientific claim, automatically organizing and combining insights from academic literature, and globally traversing descriptive mental models to identify culture-specific, causally-potent concepts and purposes.

\section*{Acknowledgments}

This material is based upon work supported by the Defense Advanced Research Projects Agency (DARPA) and Army Research Office (ARO) under Contract No. W911NF-20-C-0002 and W911NF-21-C-0007-04. Any opinions, findings and conclusions or recommendations expressed in this material are those of the author(s) and do not necessarily reflect the views of the Defense Advanced Research Projects Agency (DARPA) and Army Research Office (ARO). We thank the reviewers for their helpful feedback.

{\parindent -10pt\leftskip 10pt\noindent
\bibliographystyle{cogsysapa}
\bibliography{main}

}

\end{document}